%% file: 0_main.tex
\documentclass[sigconf,9pt]{acmart}
\AtBeginDocument{%
  \providecommand\BibTeX{{%
    \normalfont B\kern-0.5em{\scshape i\kern-0.25em b}\kern-0.8em\TeX}}}

\setcopyright{acmlicensed}
\copyrightyear{2025}
\acmYear{2025}
\acmDOI{10.1145/3658617.3697616}

\acmConference[ASP-DAC'25]{Asia and South Pacific Design Automation Conference
}{January 20--23, 2025}{Tokyo Odaiba Miraikan, Japan}
\acmISBN{979-8-4007-0635-6/25/01}

\usepackage{amsmath,amsfonts}
\usepackage{graphicx}
\usepackage{textcomp}
\usepackage{bbding}
\usepackage{pifont}
\usepackage{multicol}

\usepackage{booktabs}
\usepackage[binary-units=true]{siunitx}
\usepackage{array,multirow}
\usepackage{hyperref}
\usepackage{amsmath}
\usepackage{enumitem}
\usepackage{graphicx}
\usepackage{float}
\usepackage[symbol, hang,flushmargin]{footmisc}
\usepackage{bm}
\usepackage{subfigure}

\usepackage{tablefootnote}

\usepackage[noend]{algpseudocode}

\usepackage[ruled,linesnumbered]{algorithm2e}
\SetKwComment{Comment}{$\triangleright$\ }{}
\newlength\mylenin

\let\oldnl\nl 
\newcommand{\nonl}{\renewcommand{\nl}{\let\nl\oldnl}} 

\newlength\mylenout

\newcommand{\figref}[1]{Figure~\ref{#1}}
\newcommand{\secref}[1]{Section~\ref{#1}}

\newcommand{\tabref}[1]{Table~\ref{#1}}

\newcolumntype{L}[1]{>{\raggedright\let\newline\\\arraybackslash\hspace{0pt}}m{#1}}
\newcolumntype{C}[1]{>{\centering\let\newline\\\arraybackslash\hspace{0pt}}m{#1}}
\newcolumntype{R}[1]{>{\raggedleft\let\newline\\\arraybackslash\hspace{0pt}}m{#1}}

\usepackage{soul}
\usepackage{xurl}
\usepackage{xspace}
\usepackage{tablefootnote}

\begin{document}
\settopmatter{printacmref=false} 
\title{Exploring Code Language Models for Automated HLS-based Hardware Generation: Benchmark, Infrastructure and Analysis}


\author{
Jiahao Gai$^{2}$, Hao (Mark) Chen$^{1}$, Zhican Wang$^3$, Hongyu Zhou$^4$, \\ Wanru Zhao$^2$, Nicholas Lane $^2$, Hongxiang Fan$^{1\&2}$\\
{\normalsize $^1$Imperial College London, $^2$University of Cambridge, $^3$Shanghai Jiao Tong University, $^4$University of Sydney}\\ 
{\normalsize Email: jg2123@cam.ac.uk, hongxiangfan@ieee.org}\\
}
\renewcommand{\shortauthors}{Gai et al.}
\renewcommand{\shorttitle}{Exploring Code Language Models for Automated HLS-based
Hardware Generation}
\begin{abstract}
Recent advances in code generation have illuminated the potential of employing large language models (LLMs) for general-purpose programming languages such as \textit{Python} and \textit{C++}, opening new opportunities for automating software development and enhancing programmer productivity. 
The potential of LLMs in software programming has sparked significant interest in exploring automated hardware generation and automation.
Although preliminary endeavors have been made to adopt LLMs in generating hardware description languages (HDLs) such as \textit{Verilog} and \textit{SystemVerilog}, several challenges persist in this direction.
First, the volume of available HDL training data is substantially smaller compared to that for software programming languages. 
Second, the pre-trained LLMs, mainly tailored for software code, tend to produce HDL designs that are more error-prone. 
Third, the generation of HDL requires a significantly higher number of tokens compared to software programming, leading to inefficiencies in cost and energy consumption.
To tackle these challenges,
this paper explores leveraging LLMs to generate High-Level Synthesis (\textit{HLS})-based hardware design.
Although code generation for domain-specific programming languages is not new in the literature,
we aim to provide experimental results, insights, benchmarks, and evaluation infrastructure to investigate the suitability of \textit{HLS} over low-level HDLs for LLM-assisted hardware design generation.
To achieve this, we first finetune pre-trained models for \textit{HLS}-based hardware generation, using a collected dataset with text prompts and corresponding reference \textit{HLS} designs.
An LLM-assisted framework is then proposed to automate end-to-end hardware code generation, which also investigates the impact of chain-of-thought and feedback loops promoting techniques on \textit{HLS}- design generation.
Comprehensive experiments demonstrate the effectiveness of our methods.
Limited by the timeframe of this research, we plan to evaluate more advanced reasoning models in the future.
\end{abstract}

\maketitle
\textbf{\small ACM Reference Format:} \newline
{\small Jiahao Gai, Hao (Mark) Chen, Zhican Wang, Hongyu Zhou, Wanru Zhao, Nicholas Lane, Hongxiang Fan. 2025. \shorttitle. In \textit{Proceedings of Asia and South Pacific Design Automation Conference (ASP-DAC’25)}. ACM, New York,
NY, USA, 8 pages. \url{https://doi.org/10.1145/3658617.3697616}}
\input{./text/1_introduction}
\input{./text/2_background}

\input{./text/3_dataset}
\input{./text/4_framework}

\input{./text/5_evaluation}
\input{./text/6_conclusion}

\bibliographystyle{ACM-Reference-Format}
\bibliography{./text/aspdac_llmhls}

\end{document}

%% file: text/1_introduction.tex
\begin{figure}[t]
\centering
\includegraphics[width=85mm]{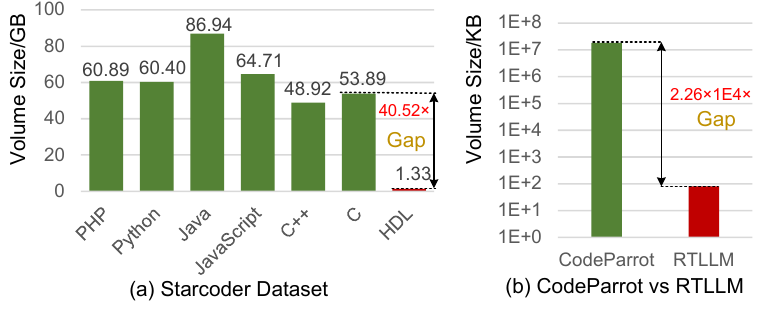}
\vspace{-7mm}
\caption{Comparison of data availability between HDLs and other software programming languages.}
\vspace{-5mm}
\label{fig:comp_data_avail}
\end{figure}

\section{Introduction}

In the field of Generative AI (GenAI), significant strides have been made in producing complex and creative content across various domains such as text~\cite{brown2020language}, image~\cite{avrahami2022blended,saharia2022image}, and video~\cite{opensora}. 
Among various GenAI technologies, large language models (LLMs) have emerged as particularly influential techniques in the realm of natural language processing~\cite{zhao2023survey}. 
This great capability of LLMs also raises intensive industrial interests in leveraging these models in automated code generation, as evidenced by GitHub Copilot~\cite{chen2021evaluating} and DeepMind’s AlphaCode~\cite{li2022competition}.
Meanwhile, over 50 pre-trained models and more than 170 programming language datasets have been published in the past few years~\cite{zhang2023unifying}.
Although significant progress has been made in this direction, most of these works mainly focus on software code generation~\footnote{This paper mainly focuses on text-to-code generation.}, and the potential of LLM for hardware design generations has not been fully exploited.

\begin{figure}[t]
\centering
\includegraphics[width=85mm]{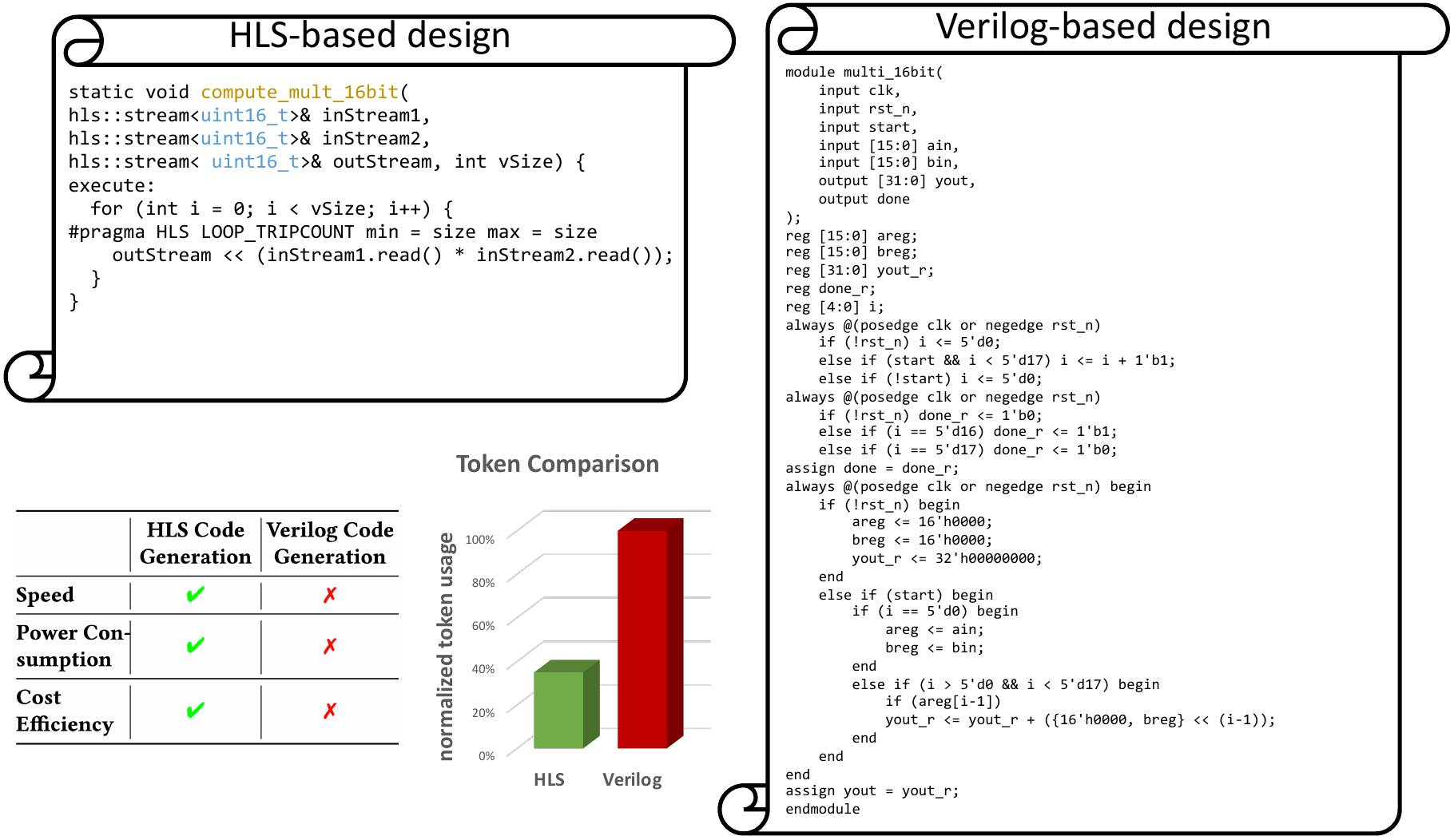}
\vspace{-5mm}
\caption{\textit{HLS}-based and \textit{Verilog}-based programs.}
\vspace{-7mm}
\label{fig:hls_vs_hdl}
\end{figure}

The promises of LLM-assisted software programming has led to several recent attempts to explore automated code generation for hardware description languages (HDLs) such as \textit{Verilog} and \textit{SystemVerilog}~\cite{li2023starcoder, lu2024rtllm, liu2023verilogeval, thakur2023verigen}.
Although multiple datasets, pre-trained models, and code infrastructures have been introduced, several key challenges reamin in LLM-assisted hardware design generation: 

\begin{itemize}[leftmargin=*]
    \item \textbf{The data availability of HLD designs for LLM training and finetuning}. \figref{fig:comp_data_avail} compares the volume of training samples for software programming languages versus HDLs.
    For instance, the general-purpose code dataset \textit{StarCoder}~\cite{li2023starcoder}, as presented in~\figref{fig:comp_data_avail} (a), shows that the available amount of HDL designs is less than $1\%$ of those for \textit{C++}.
    Similar trends can also be observed in specialized datasets, such as \textit{RTLLM}~\cite{lu2024rtllm} for \textit{Verilog} and \textit{CodeParrot}~\cite{tunstall2022natural} for \textit{Python}. \figref{fig:comp_data_avail}(b) indicates that the dataset size of \textit{RTLLM} is less than $1\%$ of that for \textit{CodeParrot}. Therefore, the quantity of training data available for hardware design is significantly lower than for software programming languages. 
    \item \textbf{Inability of utilizing learned knowledge from pre-trained coding LLMs}. Pre-trained coding LLMs are primarily trained on software programming languages, which differ significantly in semantics and syntax from HDLs. Therefore, the knowledge acquired during pre-training cannot be directly applied to hardware code generation, compounding the data scarcity issue. 

    \item \textbf{Cost of HDL generation using LLMs}. \figref{fig:hls_vs_hdl} illustrates the number of tokens required for generating identical hardware designs using High-Level Synthesis (\textit{HLS}) versus HDL. It shows that HDL implementations require approximately $3 \thicksim 4$ times more tokens than \textit{HLS} designs, making HLS-based design generation a more sustainable solution considering the latency, energy, and monetary costs associated with LLM inference. 
    
\end{itemize}

To address the aforementioned issues, 
this paper proposes an LLM-assisted framework for generating \textit{HLS}-based\footnote{This paper focues on C-based \textit{HLS}.} hardware designs.
By crawling \textit{HLS} designs from open-source Github repositories,
we collect a dataset to facilitate the fine-tuning of pre-trained LLM for the downstream \textit{HLS} code generation.
The benefits of generating \textit{HLS} code are two folds:
\textit{i)} Given that \textit{HLS} shares main semantics and syntax with \textit{C/C++}, the coding knowledge acquired during the pre-training phase of the LLMs can be effectively utilized for hardware design. 
This compatibility also reduces the learning curve and dataset requirements for fine-tuning, as the additional knowledge needed for \textit{HLS} is less than for traditional HDL coding.
\textit{ii)} The number of tokens required to generate \textit{HLS} code is lower compared to HDLs, rendering our approach more cost-effective and energy-efficient than previous methodologies.
To further improve the quality of the generated designs, the framework incorporates debugging feedback loops and a chain-of-thought enhancement mechanism, systematically integrating detected bugs back into the input for iterative refinement. 
Overall, our contributions can be summarized as follows:

\begin{itemize}[leftmargin=*]
    \item Finetuning pre-trained code language models for \textit{HLS}-based hardware generation, using a collected dataset with over $40,000$ data entries, each containing a text prompt and the corresponding \textit{HLS}-based hardware design (\secref{subsec:benchmark}).
    \item Developing a framework that automatically produces \textit{HLS} designs from input prompts, with an end-to-end evaluation of the syntax and functionality correctness (\secref{subsec:hwgen_overview}). 
    \item Integrating multiple optimization techniques such as feedback loops and chain-of-thought techniques, which improve the pass rate for syntax and functionality (\secref{subsec:hwgen_cot} \& \secref{subsec:hwgen_loop}).
\end{itemize}

%% file: text/2_background.tex
\section{Background and Related work}
\subsection{LLM-Assisted Software Engineering}\label{sebsec:background_llm_software}

Based on the modality of inputs and outputs,
language models for software engineering can be categorized into several downstream code tasks~\cite{zhang2023unifying} such as text-to-code (code generation/synthesis~\cite{ling2016latent}), code-to-text (code summarization~\cite{iyer2016summarizing}), and code-to-pattern processing (defect detection~\cite{ray2016naturalness}).
This paper focuses on code generation that aims at producing code from natural language descriptions/prompts.
To facilitate the development of code generation with language models, various datasets, approaches, and pre-trained models have been introduced over the past decade.

Due to the lack of model capability,
the early-stage methods~\cite{ling2016latent, iyer2018mapping} of code generation mainly focus on a few programming languages such as \textit{Python} or \textit{Java}.
Subsequently,
with the increasing computational power,
larger datasets are introduced to train models for multiple general-purpose programming languages.
\textit{CodeXGLUE}~\cite{lu2021codexglue} presents a comprehensive code dataset consisting of different code tasks such as clone detection and code repair.
\textit{HumanEval}~\cite{chen2021evaluating} dataset together with the code model \textit{CodeX} marks as a milestone by using pre-trained LLMs for code generation.
The promising capability shown by \textit{CodeX} sparks significant academic and industrial interests in developing LLM-assisted code generation. 
Larger code datasets, such as \textit{StarCoder}~\cite{li2023starcoder} and \textit{CodeParrot}~\cite{tunstall2022natural}, and LLMs, including \textit{Code-LLaMA}~\cite{roziere2023code} and \textit{CodeFuse}~\cite{di2023codefuse}, are open-sourced in this community.
However, most of these recent efforts focus on software programming languages.

\subsection{Automated Hardware Design Generation}\label{sebsec:background_hwgen}

Building on the success of LLM-assisted software programming, 
recent studies have explored using language models for automated hardware generation.
Since the data is the key to training LLMs for hardware code generation,
multiple HDL datasets have been introduced recently.
Thakur~\textit{et al.} present \textit{Verigen}~\cite{thakur2023verigen} dataset that contains $17$ hardware designs with $0.3$K lines of HDL code.
To increase the diversity of hardware designs for training and evaluation,
Lu~\textit{et al.} open-source a larger benchmark consisting of $30$ designs with more than $2.5$K lines.
Sourced from \textit{HDLBits}\footnote{\url{https://hdlbits.01xz.net/wiki/Main_Page}},
\textit{Verilogeval}~\cite{liu2023verilogeval} introduces larger datasets with $156$ problems.
These open-sourced datasets provide public benchmarks for text-to-HDL generation.

The evaluation metrics of LLM-assisted hardware code generation focus on three aspects: \textit{i)} syntax, \textit{ii)} functionality, and \textit{iii)} quality.
In previous literature~\cite{chen2021evaluating,liu2023verilogeval}, syntax correctness and functionality are measured using pass$@k$ metric which represents whether any of $k$ generated code samples can pass the syntax check of synthesis tools or functional unit tests.
The quality usually is defined as power, performance, and area of the generated hardware, collectively reflect the capability of the code generation methods.

Aiming at improving these metrics,
existing approaches~\cite{thakur2023verigen, lu2024rtllm, liu2023verilogeval} fine-tune pre-trained LLMs on the downstream task with optimized sampling schemes. 
These LLMs are mainly pre-trained using software programming languages.
\textit{RTLFixer}~\cite{tsai2023rtlfixer} introduces an automated framework that adopts retrieval-augmented generation~\cite{lewis2020retrieval} and ReAct prompting~\cite{yao2022react} to enable LLM-assisted debugging for RTL designs.
\textit{LLM-VeriPPA}~\cite{llm-verippa} enhances the code generation of RTL using a two-stage refinement process to progressively improve syntax, functionality, and hardware performance.
However, these approaches focus on low-level hardware languages instead of HLS. In this work, we take the first step to investigate the HLS code generation with LLM. 
Since \textit{HLS} shares similar semantics and syntax with programming languages commonly used during LLM pre-training,
this paper explores whether \textit{HLS} is better than low-level hardware languages for automated hardware design generation.
Although code generation with feedback and CoT prompting is not new in the literature of coding language models,
our experimental results, insights, benchmark, and evaluation infrastructure specific to LLM-assisted \textit{HLS} design offer valuable contributions to the future development of automated hardware generation.

%% file: text/3_dataset.tex
\section{HLS Generation Benchmark}
\label{subsec:benchmark}

\subsection{Format of Design Points}\label{subsec:benchmark_format}
\begin{figure}[bt]
\centering
\includegraphics[width=60mm]{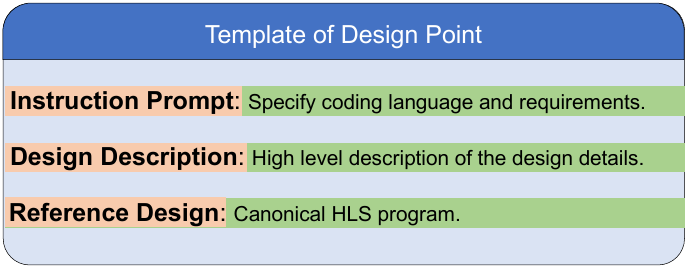}
\vspace{-2mm}
\caption{Template of design points.}
\vspace{-5mm}
\label{fig:dataset_template}
\end{figure}

Following the practices of \textit{Python} benchmark HumanEval~\cite{chen2021evaluating} and \textit{Verilog} dataset VerilogEval~\cite{liu2023verilogeval},
each design point has three components: \textit{i)} user instruction prompts, \textit{ii)} design descriptions and \textit{iii)} reference designs. 
\figref{fig:dataset_template} shows the standardized format template,  with each data point stored as a \textit{JSONL} following the Alpaca format.
The default user instruction prompt is \textit{Generate HLS code with the following instructions:}, which can be enhanced using the chain-of-thought (COT) prompting technique as detailed in ~\secref{subsec:hwgen_cot}.

\subsection{Dataset Collection}\label{subsec:benchmark_dataset}

We collect $52$ \textit{HLS}-based designs from open-source repositories, including \textit{HLSyn}~\cite{bai2023towards}\footnote{\url{https://github.com/UCLA-DM/HLSyn}} and \textit{ML4Accel}\footnote{\url{https://github.com/UT-LCA/ML4Accel-Dataset}}.
These designs are split into training and testing sets at a 4:1 ratio and fall into five categories:
\begin{itemize}[leftmargin=*]
    \item \textbf{Matrix and Linear Algebra Operations:} Includes sparse matrix-vector multiplications, dense matrix-matrix multiplication, array transformation and stencil computations.
    \item \textbf{Scientific Simulations:} Methods for solving physical and mathematical problems such as heat distribution and electromagnetic simulations.
    \item \textbf{Statistical Computations:} Calculations of statistical metrics from datasets.
    \item \textbf{Iterative Methods:} Techniques for solving equations using iterative approaches.
    \item \textbf{Other Computational Kernels:} Specialized computational tasks like molecular dynamics and interactions, encryption algorithms, and optical flow.
\end{itemize}
Each of these $52$ designs is associated with different combinations of programs such as \texttt{PIPELINE}, \texttt{PARALLEL} and \texttt{TILE}.
We filter out the \textit{HLS} programs that are invalid,
resulting in a collection of over $42,000$ \textit{HLS} programs. The whole dataset is split into training and test sets for supervised fine-tuning and evaluation, respectively.

\subsection{Generation of Design Description}\label{subsec:benchmark_desciption}
Given that the dataset encompasses over 42,000 \textit{HLS} programs, manually generating design descriptions for each program is both labor-intensive and time-consuming. 
Inspired by both HumanEval~\cite{chen2021evaluating} and \textit{Verilog},
we utilize \textit{ChatGPT} (version $3.5$ and $4$) to automate the creation of design descriptions for the datasets. 
We append each \textit{HLS} program with this base prompt when utilizing \textit{ChatGPT} to generate the corresponding design descriptions.
Both the reference design and its generated description are stored in \textit{JSON} format, adhering to the structure outlined in~\secref{subsec:benchmark_format}. This method ensures streamlined and consistent documentation of design descriptions across the dataset.
Following the practice of~\cite{liu2023verilogeval}, we provide two versions of prompts for each \textit{HLS} program in the test set: \textit{MachineGen} and \textit{HumanRefine}.
The \textit{MachineGen} version comprises prompts and instructions generated by GPT-based models without human modifications. In contrast, the \textit{HumanRefine} includes manually refined prompts to ensure more concise and human-like instructions.

\subsection{Assessment Infrastructure}\label{subsec:benchmark_syntax}

We provide evaluation infrastructure for both syntax and functionality.
For syntax verification, we use the GCC compiler with the "\texttt{-fsyntax-only}" option. It allows us to verify the syntax without the overhead of compiling the code, thereby enhancing time and space efficiency.
Regarding functionality correctness, we design unit tests tailored for each example in the test dataset. Each test case is associated with its corresponding `source\_file`. To achieve this goal, we modified the original test JSONL file to add another attribute `source\_file`. 
The testing process involves executing both the generated code and the original source code to compare their outputs. 
For instance, if the outputs from both codes consist of matrices, we conduct a targeted comparison.
This is done by selecting specific positions within the matrices from both outputs at random and verifying if they match.

%% file: text/4_framework.tex
\begin{figure*}[htb]
\centering
\includegraphics[width=125mm]{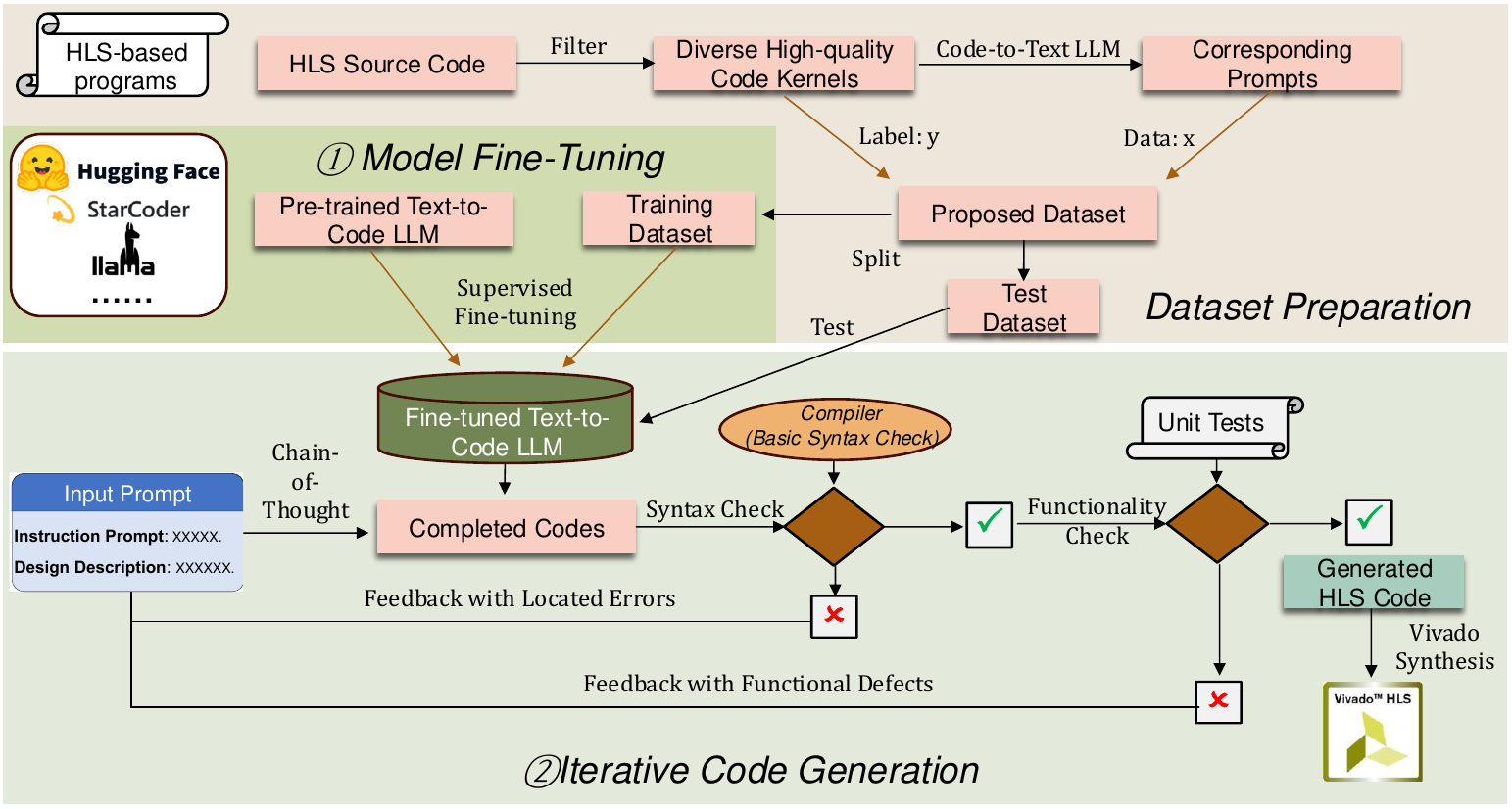}
\vspace{-3mm}
\caption{An overview of our proposed framework.}
\vspace{-3mm}
\label{fig:overview}
\end{figure*}

\section{Automated Hardware Generation}\label{subsec:hwgen}

\subsection{Framework Overview}\label{subsec:hwgen_overview}
An overview of our proposed framework is depicted in~\figref{fig:overview}.
The framework comprises two main stages: \textit{i)} model fine-tuning and \textit{ii)} iterative code generation. The final output is an HLS-based program that can be synthesized into the corresponding hardware design. While focused on \textit{Vivado-HLS}, our framework is adaptable to any HLS language with appropriate datasets.

In the model fine-tuning stage, our framework initiates by retrieving coding LLMs from open-source repositories, such as \textit{Code-Llama}\footnote{https://huggingface.co/codellama} and \textit{Start-Coder}\footnote{https://huggingface.co/blog/starcoder}.
Then, supervised fine-tuning is conducted on these pre-trained models using the HLS training data~(\secref{subsec:benchmark_dataset}).
We adopt \textit{axolotl}\footnote{https://github.com/OpenAccess-AI-Collective/axolotl} to perform the fine-tuning, allowing for customization of models and training parameters, such as learning rate, batch size, and epoch number to fit specific scenarios and available resources.

In the second stage,
we employ the fine-tuned LLM for iterative code generation.
The process begins with initial inputs consisting of user instruction prompts and design descriptions. To enhance the quality of the generated \textit{HLS} designs, we incorporate a chain-of-thought optimization technique~(\secref{subsec:hwgen_cot}) into the instruction prompts.
The code generation then proceeds with a feedback loop (\secref{subsec:hwgen_loop}) that iteratively improves the correctness of the \textit{HLS} designs.
This iterative process continues until a refined \textit{HLS} program is generated as the final output. 
Users can specify the number of iterations, providing the flexibility to navigate this trade-off according to their specific needs, with more iterations typically yielding higher quality at increased runtime expense.

\begin{figure}[t]
\centering
\includegraphics[width=70mm]{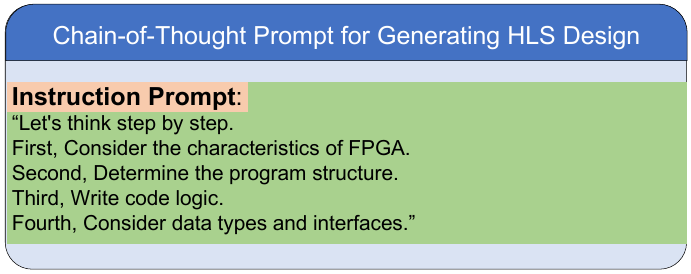}
\vspace{-3mm}
\caption{Chain-of-thought prompts for HLS generation.}
\vspace{-3mm}
\label{fig:cot}
\end{figure}

\subsection{Chain-of-Thought}\label{subsec:hwgen_cot}
Previous studies indicate that the quality of generated content is significantly influenced by the instructional prompt~\cite{zhao2021calibrate}.
The Chain-of-Thought (CoT)~\cite{wei2022chain} technique has proven simple and effective for enhancing the performance of LLMs across a wide range of tasks, including arithmetic, commonsense, and symbolic reasoning~\cite{wei2022chain}.
Although initially, COT yielded a modest $0.82$ point increase in the pass@1 metric in code generation,  this improvement was substantially augmented through structured prompting~\cite{li2023structured}.
In this paper, we investigate the effect of CoT in generating \textit{HLS}-based hardware designs. 

\figref{fig:cot} illustrates the CoT prompt structured for HLS code generation. The prompt guides a systematic approach through several targeted steps:  understanding FPGA characteristics, defining program structure, developing logic,  selecting data types and interfaces, before finalizing the code. This structured process ensures thorough consideration of each key aspect to optimize the HLS code generation.

\subsection{Two-Step Feedback Loops}\label{subsec:hwgen_loop}

Code generation with feedback loop has shown promising results in previous literature~\cite{liu2023rltf, shojaee2023execution, wang2022compilable, llm-verippa}.
This paper investigates its impact on \textit{HLS} code generation using a two-step feedback loop tailored for automated hardware generation, focusing on \textit{HLS}-related feedback.
At each iteration, the framework evaluates the syntax and functional correctness of the generated \textit{HLS} program. 
The located syntax error and functional defects are then fed back into the input prompts as additional information for subsequent code generation.

In the first step,  syntax feedback is provided using the GCC compiler with the `\texttt{-fsyntax-only}` option, as specified in \ref{subsec:benchmark_syntax}.
It captures the syntax errors without fully compiling the code. It allows rapid identification including the types and locations, and precise error mapping for targeted corrections. 
If the syntax check passes, our framework proceeds to the second step by executing predefined unit tests to compare the outputs of the generated and the original source code. 
Functional defects are recorded and added to the prompts for the next iteration. This two-step feedback loop continues for user-specified iterations, providing flexibility to balance the trade-off between design quality and runtime cost.

%% file: text/5_evaluation.tex
\section{Evaluation}

\subsection{Experimental Setup}\label{subsec:exp_setup}

In our evaluation, we adopt \textit{Code-Llama-7B} as the pre-trained model for fine-tuning, employing the low-rank-adaption (QLoRA)~\cite{hu2021lora, dettmers2024qlora} technique for faster training and lower memory consumption. 
Key configurations include loading the model in 8-bit, a sequence length of 4096, sample packing, and padding to sequence length. We set the warmup steps to 100, with a gradient accumulation of 4 steps, a micro-batch size of 4, and an inference batch size of 2.
For both syntax and functionality checks, we measure pass@3 accuracy as metrics. In the ablation study from~\secref{subsec:exp_finetune} to~\secref{subsec:complexity}, we adopt \textit{MachineGen} for evaluation.

Experiments are conducted on a server with four NVIDIA L20 GPUs (48 GB each), an 80 vCPU Intel® Xeon® Platinum 8457C, and 100GB of RAM. This setup ensures sufficient computational power and memory to handle the intensive demands of fine-tuning and inference efficiently, especially for long data sequences in the feedback loop experiment. 

\subsection{Effect of Supervised Finetuning}\label{subsec:exp_finetune}
Our first ablation study investigates the effect of the model fine-tuning.
We evaluated the performance based on both syntax and functionality checks. 
As shown in~\figref{fig:finetune_cot}(a), the results demonstrate that the finetuning dramatically increases syntax correctness from $54.85$\% to $88.44$\%. 
More importantly, the impact of finetuning is even more pronounced in the functionality evaluation, where the non-finetuned model failed to achieve any correct functionality test, but the accuracy is improved to $53.20$\% in the finetuned model. These enhancements highlight the critical role of finetuning in producing not only syntactically correct but also functionally viable codes, which demonstrates the benefits of finetuning LLMs for hardware design in the HLS code generation task.

\subsection{Effect of Chain-of-Thought Prompting}\label{subsec:exp_cot}
\begin{figure}[t]
    \centering
    \includegraphics[width=1\linewidth]{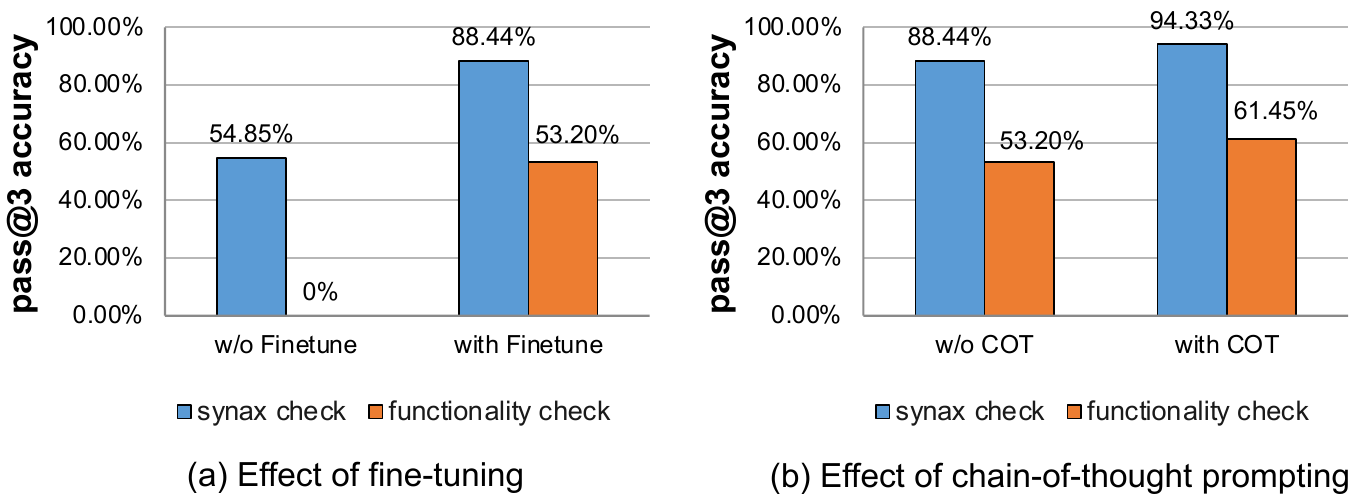}
    \vspace{-5mm}
    \caption{Effect of fine-tuning and chain-of-thought.}
    \label{fig:finetune_cot}
\end{figure}

To assess the effect of the chain-of-thought (CoT) technique,
we perform both syntax and functionality evaluation on the fine-tuned model with and without the use of CoT.
As indicated in~\figref{fig:finetune_cot}(b), incorporating CoT leads to a noticeable improvement in both metrics. 
Specifically, syntax correctness increases from $88.44$\% to $94.33$\%, and functionality score rises from $53.20$\% to $61.45$\%. 
The result demonstrates the effectiveness of CoT in enhancing the reasoning capability, thereby improving its overall performance.

\subsection{Effect of Feedback Loops}\label{subsec:exp_feedback}
\begin{figure}[t]
    \centering
    \includegraphics[width=0.95\linewidth]{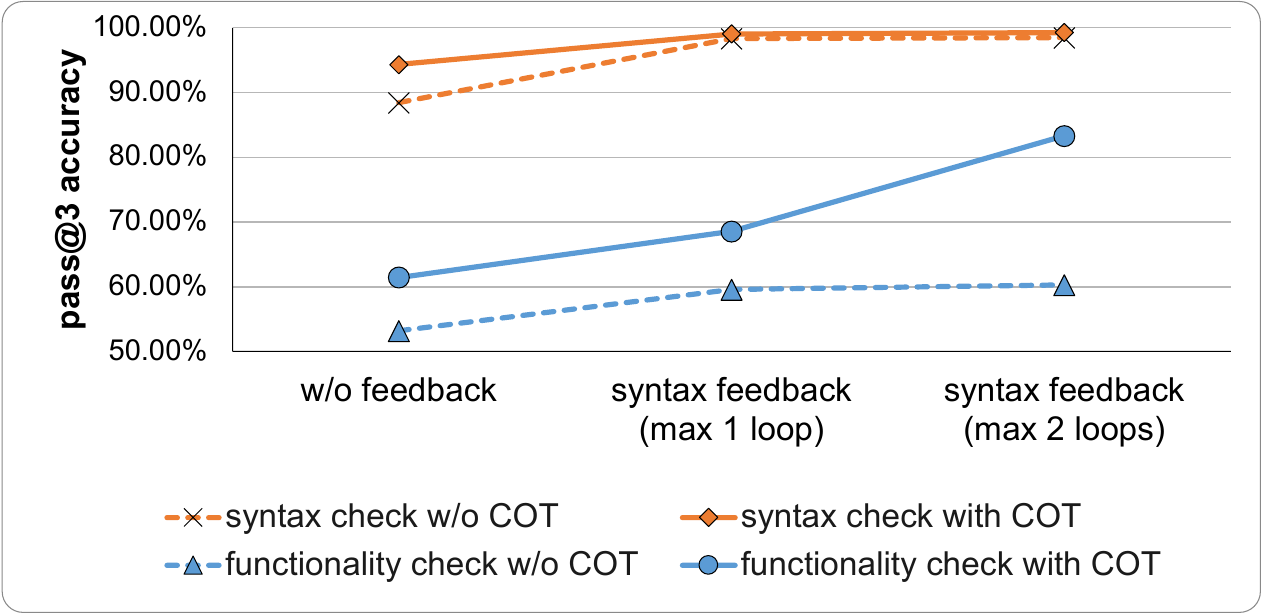}
    \vspace{-2mm}
    \caption{Effect of syntax feedback loop.}
    \vspace{-2mm}
    \label{fig:syntax_feedback}
\end{figure}

\begin{figure}[t]
    \centering
    \includegraphics[width=0.95\linewidth]{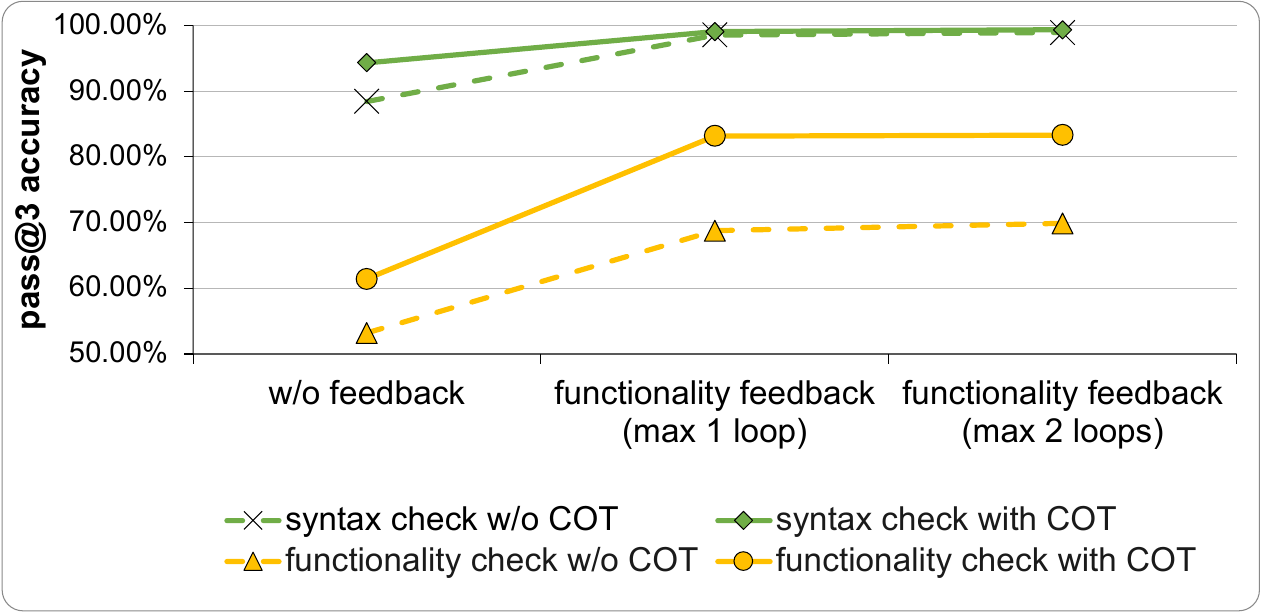}
    \caption{Effect of functionality feedback loop.}
    \vspace{-2mm}
    \label{fig:func_feedback}
\end{figure}

Our two-step feedback loop provides both syntax and functionality feedback. We evaluate the impact of these feedback loops with different numbers of iterations, ranging from 0 to 2.The results, shown in Figure ~\figref{fig:syntax_feedback} and ~\figref{fig:func_feedback}, indicate that both syntax and functionality feedback loops significantly improve model performance, especially when combined with COT prompting. The initial feedback loop yields substantial accuracy improvements in both syntax correctness and functionality evaluation, though the second loop shows diminishing returns.Syntax feedback loops enhance both syntax correctness and functionality performance, suggesting that iterative refinement is particularly effective for complex tasks. Similarly, functionality feedback loops not only improve functionality checks but also boost syntax accuracy, indicating that enhancements in functional understanding contribute to better syntactic performance.


\subsection{Time Cost and Hardware Performance}\label{subsec:exp_timecost}

\begin{figure}
    \centering
    \includegraphics[width=0.95\linewidth]{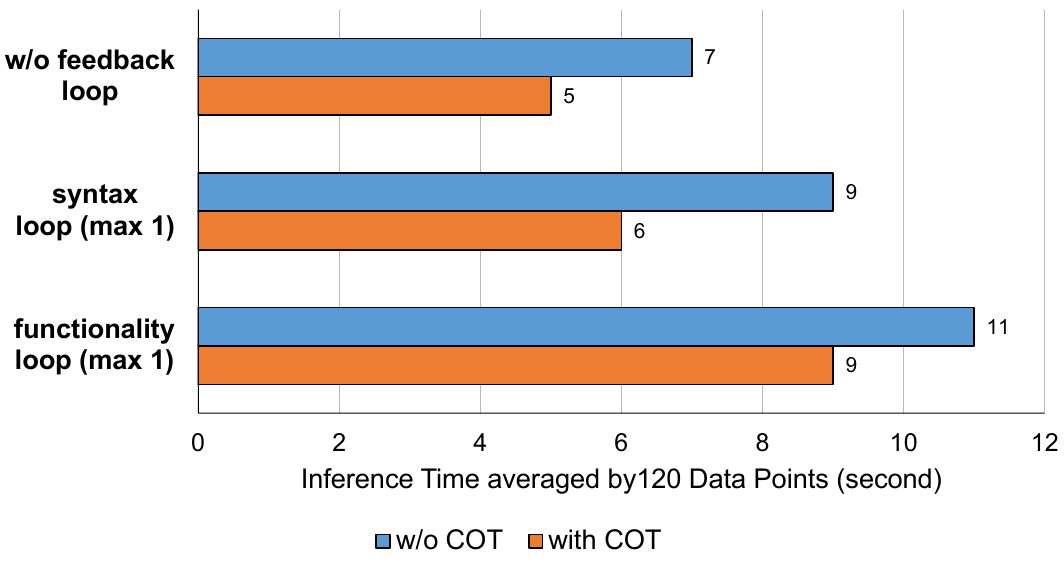}
    \vspace{-2mm}
    \caption{Time cost of code generation.}
    \label{fig:time_cost}
\end{figure}

\figref{fig:time_cost} shows the time cost for generating 120 data entries under different conditions, measuring the impact of CoT and feedback loops. Without a feedback loop, CoT significantly reduces the time. Adding a syntax feedback loop increases the time, but CoT continues to notably decrease the duration. The functionality feedback loop is the most time-consuming, though CoT still provides a notable reduction, albeit less dramatic. This demonstrates CoT's effectiveness in reducing operational times across varying complexities.

For the test set,
we evaluate the latency and resource consumption of the generated \textit{HLS} designs using a Xilinx VCU118 as our target FPGA, with a clock frequency of $200$MHz and Xilinx Vivado 2020.1 for synthesis.
As shown in~\tabref{tb:perf_resource}, all \textit{HLS} designs demonstrate reasonable performance, with BRAM usage consistently remained at zero due to the design scale.

\begin{table}[htb]
\centering
\caption{Latency and resource usage of LLM-generated designs synthesized on a VCU118 FPGA.}
\label{tb:perf_resource}
\setlength\tabcolsep{1pt} 
\scalebox{0.8}{
\begin{tabular}{C{2.5cm}|C{1.9cm}|C{1.5cm}|C{1.5cm}|C{1.3cm}|C{1.3cm}}
\toprule
{}& \textbf{Latency} (ms)& \textbf{LUTs} & \textbf{Registers} & \textbf{DSP48s} & \textbf{BRAMs} \\ \midrule
{\textbf{Available}} & - & 1182240 & 2364480 & 6840 & 4320 \\ \midrule
{\textit{ellpack}} & 0.304 & 1011 & 1079 & 11 & 0 \\
{\textit{syrk}} & 21.537 & 1371 & 1621 & 19 & 0 \\
{\textit{syr2k}} & 40.626 & 1572 & 1771 & 19 & 0 \\
{\textit{stencil2d}} & 1.368 & 287 & 123 & 3 & 0 \\
{\textit{trmm-opt}} & 15.889 & 1262 & 1239 & 11 & 0 \\
{\textit{stencil3d}} & 21.537 & 1173 & 1271 & 20 & 0 \\
{\textit{symm}} & 24.601 & 1495 & 1777 & 19 & 0 \\
{\textit{symm-opt}} & 16.153 & 1361 & 1608 & 19 & 0 \\
{\textit{symm-opt-medium}} & 579.0 & 2223 & 2245 & 22 & 0 \\
\bottomrule
\end{tabular}}
\end{table}

\subsection{Effect of Task Complexity}\label{subsec:complexity}
We analyze the effects of code complexity on the performance of fine-tuning our language model with CoT prompting and tested without the use of any feedback loops during inference. 
We categorize \textit{MachineGen} into three classes according to their code complexity: easy, medium, and difficult.
The results shown in the \tabref{tab:model_performance} indicates a clear trend: as the complexity of the generated code increases, both syntax and functionality correctness rates decline. This outcome could be attributed to several factors. First, more complex code inherently presents more challenges in maintaining syntactic integrity and functional accuracy. Second, the absence of feedback loops in the inference phase may have limited the model's ability to self-correct emerging errors in more complicated code generations.

\begin{table}[h]
\centering
\caption{Performance across different complexity levels.}
\scalebox{0.8}{
\begin{tabular}{c|c|c}
\hline
\textbf{Test Set} & \textbf{Syntax Check} & \textbf{Functionality} \\
\hline
Easy & 96.67\% & 63.33\% \\
Medium & 96.67\% & 53.33\% \\
Difficult & 90\% & 53.33\% \\
\hline
\end{tabular}}
\label{tab:model_performance}
\end{table}

\subsection{Analysis of \textit{MachineGen} and \textit{HumanRefine}}
\begin{table}[h]
\centering
\caption{Performance on \textit{MachineGen} and \textit{HumanRefine}.}
\scalebox{0.9}{
\begin{tabular}{c|c|c}
\hline
\textbf{Test Set} & \textbf{Syntax Check} & \textbf{Functionality Check} \\
\hline
\textit{MachineGen} & 93.83\% & 62.24\% \\
\hline
\textit{HumanRefine} & 47.29\% & 21.36\% \\
\hline
\end{tabular}}
\vspace{-3mm}
\label{table:eval_comparison}
\end{table}

As shown in~\tabref{table:eval_comparison}, this section compares the performance of our model on \textit{MachineGen} and \textit{HumanRefine} test sets.
Our findings reveal that the performance on the \textit{HumanRefine} is significantly lower than on the \textit{MachineGen}. This disparity suggests that the model is more adept at handling machine-generated prompts. The primary reasons for this are: the model's training data bias towards machine-generated prompts, the increased complexity and nuanced nature of human-generated prompts, and the conciseness and clarity of human-generated prompts that often omit repetitive or explicit details found in machine-generated prompts, making it harder for the model to generate syntactically and functionally correct code.

\subsection{Thoughts, Insights, and Limitations}

\noindent \textbf{1. \textit{HLS} versus \textit{HDL} for AI-assisted code generation:} The selection of programming language for hardware code generation should mainly depend on two  factors:
\begin{itemize}[leftmargin=*]
    \item \textit{Quality of Generated Hardware Design}: The evaluation of hardware design's quality includes syntax correctness, functionality, and hardware performance.
    Since \textit{HLS} shares similar semantics and syntax with programming languages commonly used during LLM pre-training, this work demonstrates that the LLM-assisted code generation for \textit{HLS} has the potential to achieve high syntax and functional correctness in hardware designs. While this work does not leverage hardware performance as feedback for design generation, it identifies this aspect as a key direction for future research and enhancements.
    \item \textit{Runtime Cost of Hardware Generation}: Although \textit{HLS}-based designs typically require fewer tokens compared to \textit{HDL} during the code generation phase—suggesting potentially lower costs—the overall runtime costs associated with HLS synthesis must also be considered. A more comprehensive quantitative comparison of these runtime costs is planned for our future work. 
\end{itemize}

\noindent \textbf{2. Input instructions and datasets are crucial}: The fine-tuning of pre-trained LLMs on \textit{HLS} dataset can bring a significant improvement in the design quality, echoing findings from previous studies on \textit{Verilog} code generation~\cite{thakur2023verigen}. 
Additionally, during our evaluation, we found that employing simple CoT prompting largely improves hardware design quality. 
This result contrasts with the application of CoT in general-purpose programming languages, where a specialized form of CoT is necessary~\cite{li2023structured}.
Therefore, future efforts for further enhancement can focus on collecting high-quality datasets and exploring better refinement of input prompts.

\noindent \textbf{3. Limitations}: At the time of this research, more advanced reasoning models, such as DeepSeek-R1~\cite{guo2025deepseek}, were not available for evaluation. Additionally, test-time scaling approaches~\cite{welleck2024decoding} could be incorporated to further enhance performance in the future.
Moreover, we observe that the diversity of hardware designs in the benchmark is limited, which may impact the generalizability of our findings.
We intend to address these limitations in our future work.

%% file: text/6_conclusion.tex
\section{Conclusion}
This paper explores automating hardware generation with code language models and  High-Level Synthesis (\textit{HLS}).
We aim to investigate the suitability of \textit{HLS} over low-level hardware
description languages for hardware design generation.
To facilitate this, we propose benchmarks and code infrastructures for evaluating LLM-assisted \textit{HLS} design generation.
Our experimental findings reveal that, with the integration of advanced optimizations such as feedback loops and chain-of-thought techniques, 
LLM-assisted \textit{HLS} code generation shows substantial promise in designing complex hardware with high levels of
syntax and functional correctness.
\newpage